\documentclass[sigconf]{aamas}  

\AtBeginDocument{%
  \providecommand\BibTeX{{%
    \normalfont B\kern-0.5em{\scshape i\kern-0.25em b}\kern-0.8em\TeX}}}

\usepackage{booktabs}
\usepackage{graphicx}
\usepackage{amsthm}
\usepackage{algorithm}
\usepackage{algpseudocode}
\usepackage{siunitx}

\theoremstyle{definition}

\newtheorem*{runexmp*}{Running Example}
\usepackage{xcolor}
\colorlet{comment}{red!40}
\colorlet{wording}{purple!40}
\colorlet{content}{blue!40}
\colorlet{question}{yellow!40}
\colorlet{info}{green!40}

\colorlet{inline-comment}{comment!200}
\colorlet{inline-wording}{wording!200}
\colorlet{inline-content}{content!200}
\colorlet{inline-question}{question!200}
\colorlet{inline-info}{info!200}

\usepackage[color=comment]{todonotes} \presetkeys{todonotes}{inline}{}

\newcommand{\ie}{i.e.\ }
\newcommand{\eg}{e.g.\ }
\newcommand{\cf}{c.f.\ }

\usepackage{flushend}
\setcopyright{ifaamas}  
\copyrightyear{2020} 
\acmYear{2020} 
\acmDOI{} 
\acmPrice{} 
\acmISBN{} 
\acmConference[AAMAS'20]{Proc.\@ of the 19th International Conference on Autonomous Agents and Multiagent Systems (AAMAS 2020)}{May 9--13, 2020}{Auckland, New Zealand}{B.~An, N.~Yorke-Smith, A.~El~Fallah~Seghrouchni, G.~Sukthankar (eds.)}  

\graphicspath{{figures/}}

\begin{document}

\title{Inference-Based Strategy Alignment for General-Sum Differential Games}

\author{Lasse Peters}
\affiliation{%
  \institution{Hamburg University of Technology}
  \streetaddress{Am Schwarzenberg-Campus 1}
  \city{Hamburg}
  \country{Germany}
  \postcode{21073}
}
\email{lasse.peters@tuhh.de}

\author{David Fridovich-Keil}
\affiliation{%
  \institution{University of California, Berkeley}
  \streetaddress{721 Sutardja Dai Hall}
  \city{Berkeley}
  \state{California}
  \country{USA}
  \postcode{94720}
}
\email{dfk@eecs.berkeley.edu}

\author{Claire J. Tomlin}
\affiliation{%
  \institution{University of California, Berkeley}
  \streetaddress{721 Sutardja Dai Hall}
  \city{Berkeley}
  \state{California}
  \country{USA}
  \postcode{94720}
}
\email{tomlin@eecs.berkeley.edu}

\author{Zachary N. Sunberg}
\affiliation{%
  \institution{University of Colorado, Boulder}
  \streetaddress{}
  \city{Boulder}
  \state{Colorado}
  \country{USA}
}
\email{zachary.sunberg@colorado.edu}

\begin{abstract}  
  In many settings where multiple agents interact, the optimal choices for each agent depend heavily on the choices of the others.
  These coupled interactions are well-described by a general-sum differential game, in which players have differing objectives, the state evolves in continuous time, and optimal play may be characterized by one of many equilibrium concepts, \eg, a Nash equilibrium.
  Often, problems admit multiple equilibria.
  From the perspective of a single agent in such a game, this multiplicity of solutions can introduce uncertainty about how other agents will behave.
  This paper proposes a general framework for resolving ambiguity between equilibria by reasoning about the equilibrium other agents are aiming for.
  We demonstrate this framework in simulations of a multi-player human-robot navigation problem that yields two main conclusions:
  First, by inferring which equilibrium humans are operating at, the robot is able to predict trajectories more accurately, and second, by discovering and aligning itself to this equilibrium the robot is able to reduce the cost for \emph{all} players.
\end{abstract}


\maketitle


\section{Introduction}

\begin{figure}
  \centering
  \includegraphics[width=0.95\linewidth]{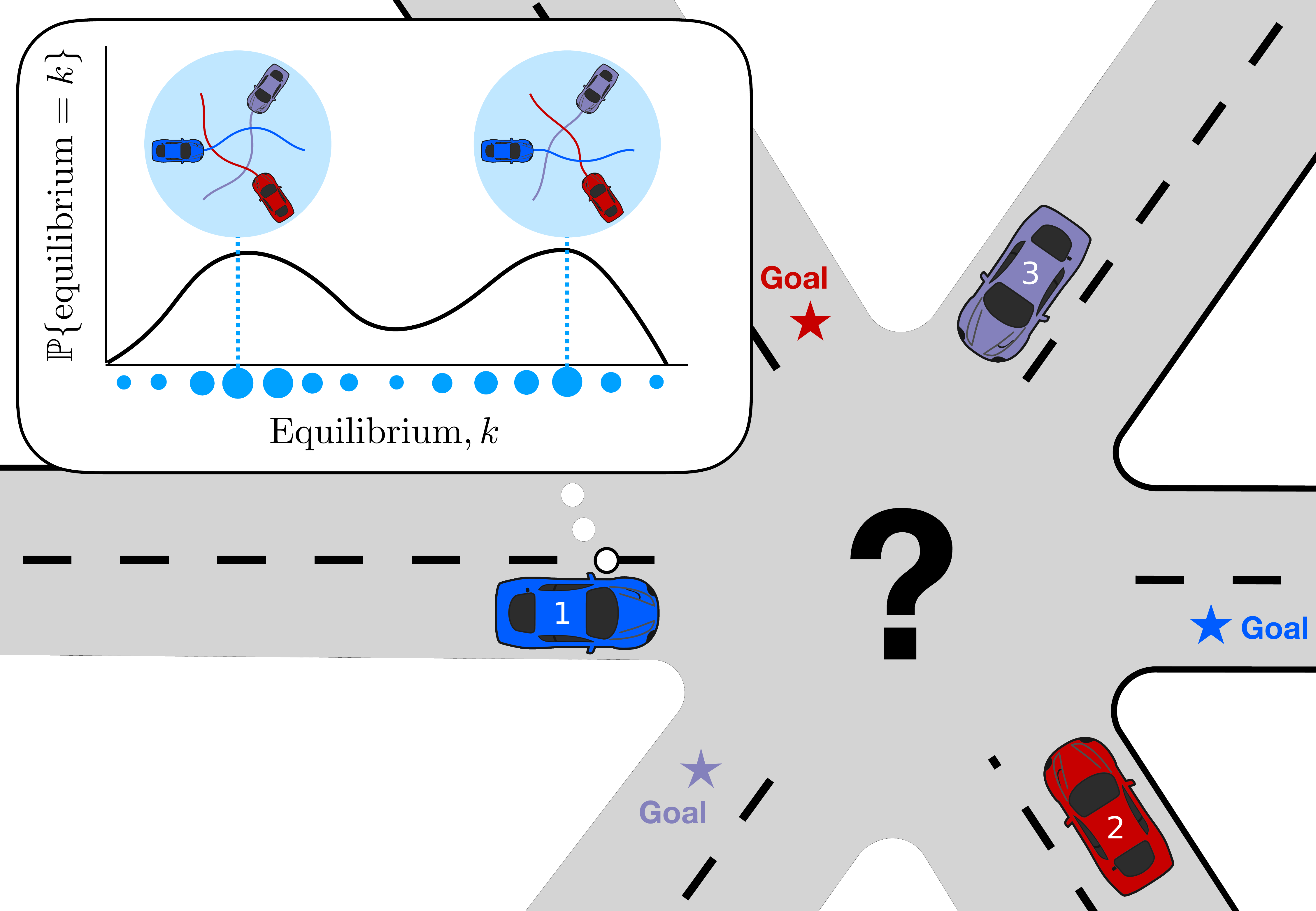}
  \caption{Illustration of the strategy alignment problem in a 3-player navigation scenario that admits multiple local equilibria.}%
  \label{fig:3-player-intro}
\end{figure}

Artificial intelligence has the potential to improve human life in many ways, for example making ground and air travel safer, automating repetitive and mundane tasks, and making knowledge and information easier to access.
One of the biggest technical challenges that has been encountered in early attempts at autonomy is consistent, safe, and efficient interaction with humans.
A common way of approaching this problem is to assume a behavior model for humans and then choose actions for the autonomous agent that minimize a cost function given this model.
The human model may be specified by hand as an open-loop prediction or a feedback policy
\cite{sunberg2017value, hubmann2018belief} or by fitting behavior to data \cite{schmerling2018multimodal, gupta2018social, ziebart2008maximum, pfeiffer2016predicting}.
However, simple models may not match the real world, and it is difficult to gather enough data or guarantee that learned models will generalize correctly to cases outside the training distribution.

An alternative to specifying a behavior model is to specify or learn a cost function and assume that other agents behave in a way that minimizes the cost they accrue.
In this setting, the optimal actions of a given agent depend heavily on the actions taken by other agents.
A mathematical formalism that captures this interaction is a general sum dynamic game.
Optimal play in such games may be characterized by one of many equilibrium concepts, \eg, a Nash equilibrium, a condition in which no player will unilaterally choose to switch strategies.
Since finding a global Nash equilibrium is generally computationally intractable \cite{conitzer2002complexity}, several techniques have been developed to find local approximations of Nash equilibria for specific problem classes \cite{mazumdar2018convergence, wang2019racing}, and recent advances have made it possible even for multi-player differential games in continuous time with continuous state and control spaces \cite{wang2019racing, wang2019driving, fridovich2019efficient, fridovich2019flat}.

However, one important drawback is that many problems of interest have multiple local equilibria.
Moreover, even if a dominating equilibrium that minimizes the cost for all players exists, expecting other agents to find this optimal solution may be overly optimistic.
If agents behave according to strategies corresponding to different equilibria, the resulting cost may be very high for all players.
Therefore, while recent work provides efficient algorithms to find approximate Nash equilibria for a wide range of problems \cite{wang2019racing, fridovich2019efficient}, it is not immediately clear how an autonomous agent should choose among multiple equilibria in practice.
We use the term \emph{strategy alignment problem} to refer to the challenges associated with the presence of multiple equilibria.

One potential solution to this issue would be to establish a strict hierarchy between agents, allowing leading players to announce their plans and expect others to follow, thus relaxing the interactive game to a sequential planning problem \cite{chen2015safe}.
However, this approach necessitates some form of direct communication between agents which is, in itself, a challenging problem as inhomogeneous groups of agents (\eg humans and robots) may struggle to explain their plans to each other \cite{langley2017explainable}.

In this work, we propose a general framework that addresses the strategy alignment problem without direct communication or negotiation between agents.
Instead of requiring globally optimal decisions of all players, we assume that other agents approximately choose their strategies according to a local equilibrium of the game.
To handle the resulting uncertainty arising from the multiplicity of local equilibria, we use a particle filtering technique, abstractly illustrated in \cref{fig:3-player-intro}, to infer the equilibrium that other players are aiming for.
By utilizing this information, we enable the autonomous agent to make more accurate predictions and align itself to this equilibrium to compute a more efficient strategy.

\section{Background \& Related Work}
To put this work in context, we provide a brief overview of related work that is concerned with modelling and planning in multi-player settings.
Here, we survey both game-theoretic approaches as well as other techniques that can be broadly categorized as partially observable Markov decision process (POMDP) approximations.

\paragraph{Game Theoretic Approaches}
Various works have approached the problem of multi-agent interaction from a game-theoretic perspective.
As the mutual dependence of each players actions on the decisions of others poses a computational challenge,
a common approach is to simplify the problem by establishing a leader-follower hierarchy between players to arrive at a Stackelberg dynamic game \cite{simaan1973stackelberg}.
Such approaches have been demonstrated in the context of human-robot interaction \cite{sadigh2016planning, sadigh2016information, yoo2012stackelberg} but have been shown to yield undesirably aggressive behavior of the leader \cite{fisac2019hierarchical}.

Other approaches avoid this pure leader-follower structure and aim for more symmetric roles of different players.
In \cite{fisac2019hierarchical}, the interaction is modelled in a hierarchical approach that solves a fully coupled dynamic game to inform a low-level controller.
However, this approach solves the high-level Nash game through discretization of the state and input space and thus does not easily scale to multiple players.

To avoid the curse of dimensionality while maintaining symmetric roles of different players, recent work has focused on applying algorithms akin to differential dynamic programming to find approximate local Nash equilibria in dynamic games \cite{fridovich2019efficient, fridovich2019flat}.
While these works make contributions in efficiently computing locally optimal solutions to general-sum differential games, they do not address the key practical challenge of choosing among multiple local equilibria.
In this paper, we directly build on the results of \cite{fridovich2019efficient, fridovich2019flat}.
Based on this local game-solver, we propose a general framework for reasoning about which equilibrium other players are operating at and align the ego agent's strategy accordingly to handle cases in which a unique solution does not exist.

\paragraph{POMDP Approximations}

Several works address behavioral uncertainty by modeling certain aspects of human behavior as latent state variables and maintaining belief over these variables to compute optimized decisions.
In the field of autonomous driving, inference of behavioral parameters has been demonstrated to provide a significant benefit when interacting with other drivers \cite{sunberg2017value} and a significant amount of work has focused on using this information in approximate POMDP schemes \cite{ward2017probabilistic, sunberg2017value, hubmann2018belief, bouton2019cooperation}.
However, these works typically use highly simplified models like IDM \cite{treiber2000congested} and MOBIL \cite{kesting2007general} for the behaviors of other players.
Similarly, in \cite{cunningham2015mpdm, galceran2015multipolicy, mehta2016autonomous} a library of hand-engineered feedback strategies is used to model the behavior of other agents.
While these work demonstrate the benefit of employing behavioral inference, the policies used to model the behavior of other agent's are somewhat arbitrarily chosen and may not specify suitable behavior for all cases.

Other works model humans as rational agents seeking to maximize their own objective function.
In \cite{fridovich2019confidence, bajcsy2019scalable}, humans actions are predicted as the outcome of a noisily-rational decision process \cite{baker2007goal} with unknown goals.
Here, inference is used to reason about both the intentions of humans as well as the accuracy of the predictive model.
However, these works treat agents as independently optimizing players and capture interaction between multiple agents only indirectly by reducing the confidence of predictions.

In this work, we model interaction explicitly by casting this problem as dynamic game and use inference to discern between different modalities that can result even with known objectives.
We stress, however, that the framework we present is more general and may in future work be used to capture other sources of uncertainty including uncertainty in the objectives of humans.

\section{Problem Statement and Approach}\label{sec:problem}

We consider a single autonomous agent $\mathcal{A}$ operating in a shared space with $N-1$ humans $\mathcal{H}_i$, and use the convention that the autonomous agent always has index $1$.
The joint state, $x$, of all agents evolves over time, $t$, according to an ordinary differential equation
\begin{equation}
  \dot{x} = f(t, x, u_1, \ldots, u_n),
\end{equation}
where each agent $i$ is in control of an input $u_i$ to the system of joint dynamics.
Additionally, each agent has a cost
\begin{equation}
  J_i(u_1, \ldots, u_N) = \int_0^T g_i(t, x, u_1, \ldots, u_N) dt,
\end{equation}
defined as an integral of running costs $g_i$ over a finite horizon $T$ and implicitly depends upon the initial state $x(0)$.
The objective of each player is to minimize their respective cost function under the constraints of the dynamics by choosing a suitable state feedback strategy $\gamma_i$, \ie
\begin{equation}
u_i(t) = \gamma_i(t, x(t)).
\end{equation}
The cost function captures the behavior of each agent by encoding different aspects of their preferences.
Together, the state dynamics and the objectives of each player pose an optimization problem that may be
cast as a finite-horizon $N$-player general-sum differential game.

\begin{runexmp*}
For the sake of clarity, we introduce a running example which we will use throughout this paper to illustrate the problem formulation as well as our approach.
In this example we consider three agents ---~a robot and two human players~--- that navigate in a shared environment with no obstacles (\cf \cref{fig:3-player-intro}).
Defining the joint state to be $x=[x_1; x_2; x_3]$, each agent's dynamics are those of a 4D unicycle
\begin{equation}
  \dot{x_i} = \begin{bmatrix}\dot{p}_{x,i}; \dot{p}_{y,i}; \dot{\theta}_i; \dot{v}_i\end{bmatrix} = \begin{bmatrix}
  v_i\cos\theta_i; v_i\sin\theta_i; \omega_i; a_i
  \end{bmatrix},
\end{equation}
where each agent is in control of the longitudinal acceleration $a_i$ as well as the steering rate $\omega_i$ of their respective systems.
The behavior of each agent is characterized by the following objectives:
\begin{enumerate}
  \item Reach a preset goal state $x_{g,i}$ within the horizon $T$.
  \item Minimize control effort.
  \item Prefer low velocities.
  \item Avoid collisions with other players.
\end{enumerate}
For each player a piecewise quadratic cost is formulated to encode these preferences.
For the control inputs $u_i$, velocity state $v_i$ and terminal state error $||x_i(T) - x_{g,i}||$ quadratic penalties are used.
The collision-avoiding interactions are modeled collaboratively with semi-quadratic
penalties on the pairwise distances between players.
As some of the aspects of this behavior (crucially collision avoidance) depend implicitly or explicitly on the joint state of at least two agents, this problem requires each player to reason about the decisions of other players.
\end{runexmp*}

\subsection{Local Equilibrium Problem}
While in general, there exist different notions of optimal play in a dynamic game, here we assume that all agents approximately optimize for a local Nash equilibrium.
Considering an overloaded notation where $J_i$ depends upon strategy profiles $(\gamma_1, \ldots, \gamma_N)$ rather than control inputs $(u_1, \ldots, u_N)$, a local Nash equilibrium is attained if each player follows a strategy $\gamma_i^\ast$ such that the strategy profile satisfies
\begin{equation*}
\begin{split}
J_1^\ast &= J_1(\gamma_1^\ast; \gamma_2^\ast;\ldots;\gamma_N^\ast) \leq J_1(\gamma_1; \gamma_2^\ast;\ldots;\gamma_N^\ast),\\
J_2^\ast &= J_2(\gamma_1^\ast; \gamma_2^\ast;\ldots;\gamma_N^\ast) \leq J_2(\gamma_1^\ast; \gamma_2;\ldots;\gamma_N^\ast),\\
         &\mathrel{\makebox[\widthof{=}]{\vdots}}\\
J_N^\ast &= J_N(\gamma_1^\ast; \gamma_2^\ast;\ldots;\gamma_N^\ast) \leq J_N(\gamma_1^\ast; \gamma_2^\ast;\ldots;\gamma_N),
\end{split}
\end{equation*}
for $\gamma_i$ in the local neighborhood of $\gamma_i^\ast$.
An in-depth characterization of local Nash equilibria can be found in \cite{ratliff2016characterization}.
Intuitively, this condition entails that no player has a unilateral incentive to make a \emph{small deviation} from his/her current strategy.
This choice of equilibrium models symmetry of information between different players in this game.

Restricting the search to local Nash equilibria is a common relaxation \cite{mazumdar2018convergence,wang2019racing,wang2019driving} to reduce the computational complexity of the otherwise intractable optimization problem \cite{conitzer2002complexity}.
However, even with this relaxation finding local Nash equilibria remains computationally challenging \cite{mazumdar2018convergence} and many results are restricted to two-player zero-sum settings \cite{mazumdar2018convergence, mazumdar2019finding} or search over open-loop trajectories instead of feedback strategies \cite{wang2019driving, wang2019racing}.
Rather, we use a recent, computationally efficient method which finds approximate local Nash equilibria by solving a sequence of linear-quadratic (LQ) games \cite{fridovich2019efficient}.
This further relaxation yields equilibria that qualitatively resemble local Nash, though in fact they are global Nash equilibria of successive LQ approximations to the original game.
Moreover, each individual LQ game affords an analytic solution via coupled Riccati equations, making the overall algorithm extremely efficient.
We use the term \emph{approximate local Nash equilibrium} to refer to this equilibrium concept.
The qualitative nature of these equilibria is discussed in \cref{sec:nash-results}.

While there is no general theoretical guarantee that any approximate local Nash equilibria exist in a given game, in practice, they are commonly present.
For example, in all of the experiments conducted in this work, these equilibria were found.
In fact, many problems of interest have multiple local equilibria.
However, it is important to understand that even if one or more local equilibria exist, there is no guarantee that any of these are globally dominant (\ie have simultaneously lower cost for all players than any other strategy profile).
Furthermore, even if a globally dominant equilibrium exists, this equilibrium may not be unique; there may be other equilibria that achieve the same cost.

\begin{runexmp*}
For our running example it is intuitively clear that it allows for multiple local equilibria due to the structure of the cost.
That is, for each pairwise encounter of two players the involved agents need to decide on which side they pass each other.
While for some cases this conflict may be trivially resolved because both players have a strong preference for the same solution, other scenarios require some form of negotiation to resolve the ambiguity.
In the next section, we will elaborate on how we resolve this conflict.
\end{runexmp*}

\subsection{Strategy Alignment Problem}\label{sec:strategy-alignment-problem}

To resolve this conflict between multiple equilibria, we slightly break the symmetry between the players in the following manner.
First, we assume that humans $\mathcal{H}_i$ automatically agree with other human players on a local solution of the game.
This agreement between humans can be thought of as the innate ability of humans to communicate through subtle cues that are difficult for robots to pick up on.
Therefore, we do not expect humans to be able to communicate with robots with similar clarity.
Instead, it is the robot's responsibility to infer the equilibrium negotiated by the humans and align to their preferred local solution.
By this means, we preserve local symmetry but allow humans to take a leading role in globally selecting an equilibrium.

In accordance with these assumptions, the local equilibrium at which human players operate is modeled as a latent state that the agent cannot directly observe.
Instead, the agent $\mathcal{A}$ receives observations of only the physical state, $x$.
The term \emph{strategy alignment} is used to refer to the process of inferring this latent state and adapting to the corresponding equilibrium strategy.

\begin{runexmp*}

\Cref{fig:3-player-intro} illustrates the strategy alignment problem for our running example.
Here, by solving the strategy alignment problem the robot allows humans to choose their preferred local solution while maintaining shared responsibility for collision avoidance.
Another perspective on this problem is the fact that this type of strategy alignment allows the robot to account for modeling  errors in the cost used to encode the human behavior.
For example, when approaching a human that comes straight at the robot, from a robot's perspective it seems reasonable to model humans as having symmetric preference for passing on either side.
However, humans may have cultural or otherwise unmodeled preferences for a specific solution.
\end{runexmp*}

\section{Inference-Based Strategy Alignment}\label{sec:solver}

We present a general framework for solving the strategy alignment problem presented in \cref{sec:problem}.
On a low level, our approach utilizes iterative LQ approximations of the game as proposed in \cite{fridovich2019efficient} to approximate local Nash equilibria of the game.
By seeding this local solver with numerous samples from an explicitly defined distribution over initial strategies, we induce an implicitly defined distribution of local equilibria.
At every time step, we then use approximate Bayesian inference ---~in an approach similar to particle filtering~--- to recursively update the belief over the latent equilibrium with observation of the physical state $x$.
Using this a-posteriori belief we extract the most likely equilibrium and align the robot's strategy accordingly.

\subsection{Finding Local Equilibria}\label{sec:ilqgames}

Our method is, for the most part, agnostic to the solver used to compute local equilibria of the game.
There are two requirements for a solver to be used in the new framework.
First, it must admit some form of persistent state that encodes a specific equilibrium to be used for inference as measurements are received.
Second, because inference must be run at near real-time planning rates, the solver needs to be sufficiently fast to allow solving the game for multiple equilibria at every planning step.
In this work we use the iterative linear-quadratic method proposed in \cite{fridovich2019efficient}.
While for a detailed presentation of this approach we direct the reader to \cite{fridovich2019efficient, fridovich2019flat}, here we only provide a brief outline of the method to show how it is embedded in our inference scheme.

On an abstract level, the approach extends the idea of the well known ILQR algorithm \cite{li2004iterative, todorov2005generalized} to the domain of $N$-player general sum differential games.
Given the initial joint state $x_0$  as well as initial strategies $\gamma_{i,0}$ for every player, the algorithm iteratively updates this strategy until convergence using the following steps:
\begin{enumerate}
  \item Simulate the joint state trajectory using the current strategies.
  \item Locally approximate the game along this trajectory by linearizing the dynamics and quadraticizing the cost.
  \item Update the strategies using the analytic solution to the LQ approximation of the game via dynamic programming.
\end{enumerate}

If this algorithm converges, the resulting strategies compose an approximate local Nash equilibrium of the game.
Additionally, as the algorithm is deterministic, given the same initial strategies $\gamma_{i,0}$ it will arrive at the same equilibrium.
Hence, the set of initial strategies can be directly used as an implicit representation of the latent equilibrium state, fulfilling the first solver requirement.
Furthermore, initial strategies can be used to warm-start the solver.
As our inference method uses the LQ game solver in a receding horizon approach, this property allows us to rapidly update solutions using the last set of converged strategies yielding the speed that fulfills the second requirement.

\subsection{Inferring the Local Equilibrium}\label{sec:inference}

\begin{figure}
  \centering
  \includegraphics[width=0.8\linewidth]{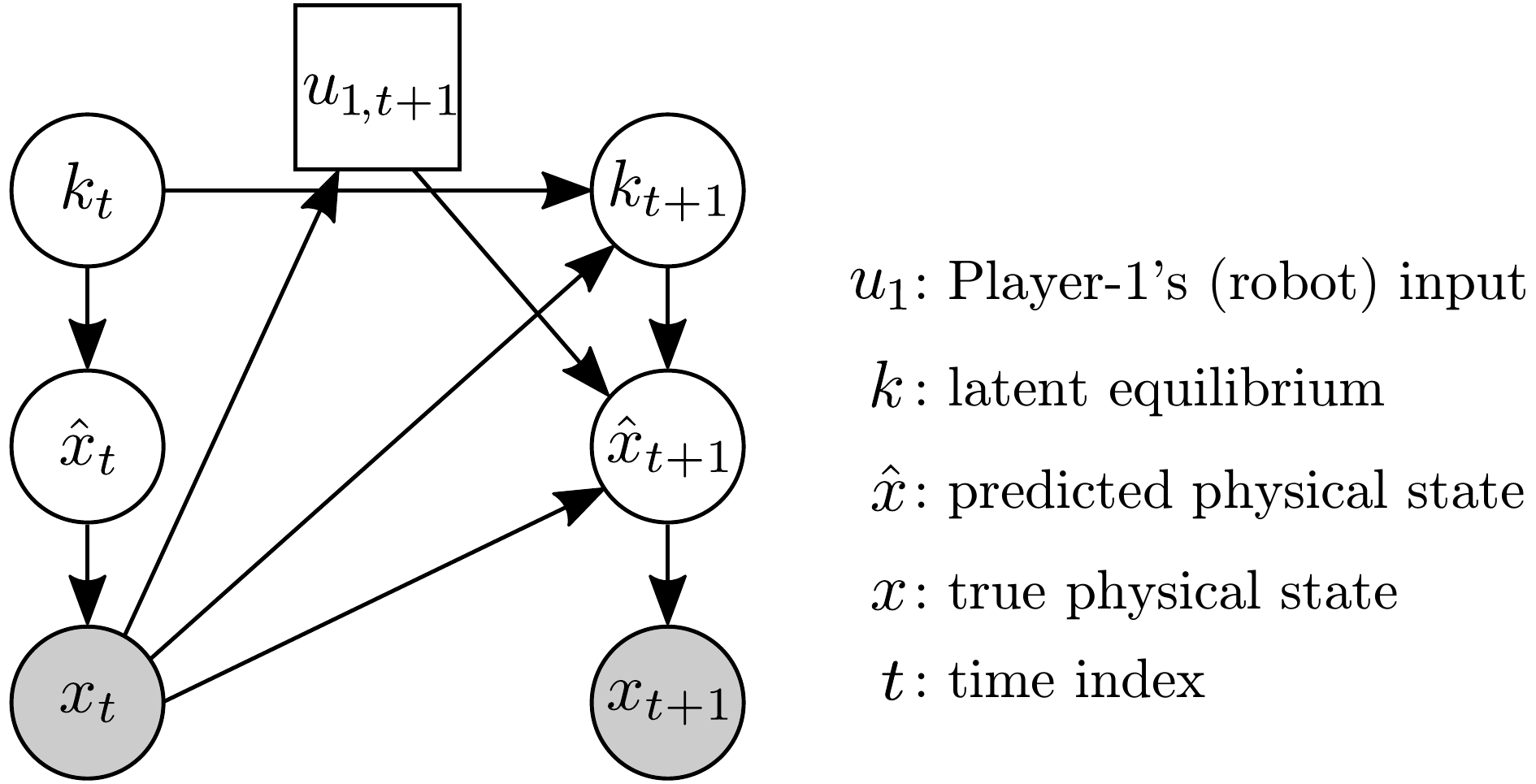}
  \caption{Dynamic decision network used to model the equilibrium inference problem.}%
  \label{fig:ddn}
\end{figure}

Since the robot does not know the equilibrium that the humans have chosen, it can only attempt to infer how likely each equilibrium is.
As we assume that the robot is able to measure the state perfectly, in a perfect mathematical abstraction, there would be a unique trajectory for each equilibrium, and the robot could eliminate solution hypotheses if they did not match the observations exactly.
However, in reality, there are several sources of uncertainty.
First, real humans do not behave exactly according to equilibrium strategies and instead follow trajectories that are difficult to model perfectly.
Second, there will always be discrepancies between the dynamics and costs in the game used for planning and the actual dynamics and preferences of humans in the real world.
Finally, there is some numerical noise in the game solutions, so even if the model dynamics and costs matched the real world perfectly, strategy profiles attained for a specific equilibrium with the method described above may deviate within the tolerance of the convergence check.

Because of this uncertainty, we use Bayesian inference to reason about the latent equilibrium.
\Cref{fig:ddn} shows the dynamic decision network used to model the corresponding inference problem.
As specified in \cref{sec:strategy-alignment-problem}, the robot's decision at time step $t+1$ is informed by past observations of the physical state, $x_{1:t}$, while the equilibrium, $k_t$, generating the human behavior remains unobserved.
At every time step, the next equilibrium, $k_{t+1}$, depends upon the previous equilibrium, $k_t$, and physical state, $x_t$, as the human players update their strategies to account for the decision of the robot.
$\hat{x}_{t+1}$ represents the predicted physical state that would be attained if all human players followed the equilibrium strategy exactly.
The true physical state, $x_{t+1}$, is modeled as an emission of $\hat{x}_{t+1}$ to account for deviations of human players from the exact equilibrium strategies due to the sources of uncertainty discussed above.
With this model, the exact Bayesian update to maintain a belief, $b$, over the latent equilibrium, $k$, may be computed as
\begin{equation}\label{eq:bayesian-update}
\begin{split}
  b_{t+1}(k_{t+1}) \propto & \int_{k_t\in\mathcal{K}}\int_{\hat{x}_{t+1}\in\mathcal{X}}
                           [ p(x_{t+1} | \hat{x}_{t+1}) \\
                           & p(k_{t+1}, \hat{x}_{t+1} | u_{t+1}, k_t, x_t) b_t(k_t)] \, d\hat{x}_{t+1} \, dk_t\text{.}
\end{split}
\end{equation}
Unfortunately, evaluating this update rule is computationally challenging as enumerating all equilibria in the equilibrium space, $\mathcal{K}$, is generally intractable \cite{conitzer2002complexity}.
However, by sampling initial strategy profiles and solving the game at these points we can sample a subset of $\mathcal{K}$.
Using this idea, we approximate the update, \cref{eq:bayesian-update}, through a particle filtering technique.

First, $K$ approximate local Nash equilibria are sampled by randomly selecting seed strategies from a problem-specific distribution and solving the game at each of these points.
This seed distribution is necessarily dependent on the specific state space and dynamics of the problem, but it was not difficult to find suitable distributions for the experiments conducted here.
In general, the seed distribution must cover the strategy space in a manner that allows to recover the relevant equilibria that human players may consider; equilibria that are not attained by any sampled seeding of the solver can not be inferred.
Therefore, the performance of our method when applied to different environments and tasks depends on the ability to specify a suitable seed distribution.
In this work, open-loop strategies, $\gamma_{i,0}(t, x) = [\beta_\omega \cos(t/T\pi); \beta_a \cos(t/T\pi)]$,
corresponding to s-shaped trajectories with uniformly sampled turn rate and acceleration parameters, $\beta_\omega$ and $\beta_a$, were used for all players.

Each of the resulting strategy profiles is referred to as a particle and has a corresponding weight.
All weights are initially 1, but as new state measurements are received at each time step, the strategies are adjusted and the weights are updated so that particles that explain the observed trajectory well have higher weights.
Specifically, when a new state measurement arrives, the game is re-solved in a receding horizon fashion (\eg warm-started using the previous game solution) to determine the updated strategies, and a prediction of the physical state, $\hat{x}$, is made with the resulting strategy profile.
From the perspective of a vanilla particle filter, this step corresponds to sampling a transition for every particle from the transition density, $p(k_{t+1}, \hat{x}_{t+1} | u_{t+1}, k_t, x_t)$ (\cf \cref{eq:bayesian-update}), where the process of solving the game and integrating the dynamics constitute the generative transition model.
The weight for the particle is then updated by evaluating the probability density $p(x | \hat{x}^{(k)})$ which captures potential deviations of humans from the exact strategy corresponding to equilibrium $k$.
These steps correspond to \cref{alg:mapplanning}, lines \ref{ln:particlefor} through \ref{ln:endparticlefor}.
Finally, after each particle has been updated, the weights for particles that represent the same equilibrium, as determined by measuring the distance between the trajectories, are combined.
This step is represented by the \textsc{CombineDuplicates} function called in line~\ref{ln:combine} of \cref{alg:mapplanning}.

The deviation model, $p(x | \hat{x})$,  is somewhat arbitrary because it is meant to capture the three difficult-to-model sources of uncertainty mentioned above.
In this work, a Gaussian distribution is used, as is commonly done in cases where uncertainty is difficult to model, but is expected to be unimodal.
Specifically for these experiments, $X\sim\mathcal{N}(\hat{x}, \epsilon_O I)$.
It would also be possible to incorporate domain knowledge into the algorithm by using a more complex distribution.

Given the widespread use and demonstrated effectiveness of particle filtering, it is likely that this technique can be extended to cover more complex scenarios including when humans switch between equilibria with techniques that improve robustness, such as resampling.

\subsection{Globally Aligned Closed-Loop Planning}

After the inference algorithm has weighted each of the sampled solutions, the robot must decide which feedback strategy to apply.
There are many possible candidates for the best strategy.
If the weights are appropriately normalized, then they define a probability distribution over the sampled equilibria particles (\ie ${P(k)=\frac{w^{(k)}}{\sum_{k\in\Gamma} w^{(k)}}}$).
This structure is analogous to a belief in a POMDP formulation \cite{kochenderfer2015decision,kaelbling1998planning} where the human player's choice of equilibrium is the latent part of the state.
Thus, solution concepts used for POMDPs are applicable here.

It is well known that to find the optimal solution of a POMDP, the agent must reason about the information they will receive in the future.
However, this is computationally intractable in general \cite{papadimitriou1987complexity}, so approximations are usually used.
Commonly used approximations include generalized QMDP \cite{sunberg2017value,littman1995learning} and planning assuming the most likely or mean latent state \cite{sunberg2017value}.
In the present setting, QMDP is difficult to apply because it is difficult to evaluate every possible control input against all of the strategies the other players might take.
Moreover, the mean latent state in this setting corresponds to the weighted average of several equilibrium strategies, which is usually not a strategy that a human would take.
For these reasons, we use only the maximum-likelihood \emph{a posteriori} equilibrium to make control decisions, specifically applying the control specified by the robot player's strategy in that equilibrium as shown in \cref{alg:mapplanning}, line~\ref{ln:control}.

\begin{algorithm}
    \caption{Maximum A Posteriori Aligned Control} \label{alg:mapplanning}
    \begin{algorithmic}[1]
        \State Sample $K$ particles $\left\{\gamma^{(k)}\right\}$ from initial strategy distribution
        \State $\Gamma \gets \left\{1..K\right\}$
        \For{each time step $t$}
            \State Receive observation $x(t)$
            \For{each particle index $k$ in $\Gamma$} \label{ln:particlefor}
                \State $\gamma^{(k)} \gets \textsc{SolveGame}\left(x(t-1), \gamma^{(k)}\right)$ \label{ln:resolve}
                \State $\hat{x}^{(k)} \gets x(t-1) + \int_{t-1}^{t}f\left(\tau, x(\tau), u_1(\tau), \gamma_{-1}^{(k)}(x(\tau)) \right) d\tau$
                \State $w^{(k)} \gets w^{(k)} p(x(t) | \hat{x}^{(k)})$ \label{ln:weightupdate}
            \EndFor \label{ln:endparticlefor}
            \State $\Gamma \gets \Call{CombineDuplicates}{\Gamma}$ \label{ln:combine}
            \State $k_\text{MAP} \gets \underset{k \in \Gamma}{\text{argmax}}\left\{w^{(k)}\right\}$
            \State $u_1(t) \gets \gamma^{(k_\text{MAP})}_1(x(t), t)$ \label{ln:control}
        \EndFor

    \end{algorithmic}
\end{algorithm}

\section{Experimental Results}

In this section, we analyze the benefits of inference-based strategy alignment by comparing our approach to a baseline approach similar to previous methods \cite{fridovich2019efficient, fridovich2019flat} that does not include inference.
For this purpose, we use the running example introduced in \cref{sec:problem}.
After stating some of the implementation details, we first demonstrate the multi-modality of the problem by enumerating some of the approximate local Nash equilibria for this problem.
We then evaluate the prediction performance of our approach by isolating it from the control part in a purely observing setting.
Finally, we study the interaction dynamics and performance of the strategy alignment approach in a closed-loop planning scenario.

\subsection{Implementation Details}\label{sec:implementation}

We implement the LQ game solver proposed in \cite{fridovich2019efficient,fridovich2019flat} as well as our strategy alignment framework using the Julia programming language\footnote{Code available at \url{github.com/lassepe/GameInference.jl}} \cite{bezanson2012julia}.
By this means, we provide a fast and flexible implementation that can be easily adapted to other scenarios and other types of uncertainty.
Our current implementation of the inference framework is single-threaded.
We stress, however, that large parts of our method are trivially parallelizable as solutions for individual particles
can be computed independently.

\Cref{tab:params} shows the parameters used throughout the experiments.

\begin{table}[tbph]
    \caption{Simulation parameters}
    \center
    \begin{tabular}{@{}llr@{}}
        \toprule
            Parameter                         & Symbol                & Value \\
        \midrule
            time step                         & $\Delta t$            & \SI{0.1}{\second} \\
            simulation horizon                & $T_s$                 & \SI{10}{\second} \\
            prediction horizon                & $T_p$                 & \SI{10}{\second} \\
            number of particles (3-player)    & $K_3$                 & \num{50} \\
            number of particles (5-player)    & $K_5$                 & \num{150} \\
            observation noise                 & $\epsilon_O$          & \num{0.1} \\
        \bottomrule
    \end{tabular}
    \label{tab:params}
\end{table}

\subsection{Local Equilibria}\label{sec:nash-results}

We begin by analyzing the qualitatively different approximate local Nash equilibria that exist in a multi-player navigation problem.
For each of the experiments presented in this section, we generate a collection of equilibria by
solving the multi-player game for randomly sampled initial strategies (\eg see \cref{fig:2-player-clustered}, right).
We then apply k-means clustering on the state trajectories traced out for each sample to discern the different types of encounter geometries.

\subsubsection{Minimal Example: 2-Player Scenario}

For the sake of clarity, we first consider a 2-player version of our running example.
This simple example demonstrates that even in the minimally complex case of an 2-player encounter a unique dominating strategy profile may not be found and thus some form of strategy alignment is required.

\begin{figure}
  \centering
  \includegraphics[width=\columnwidth]{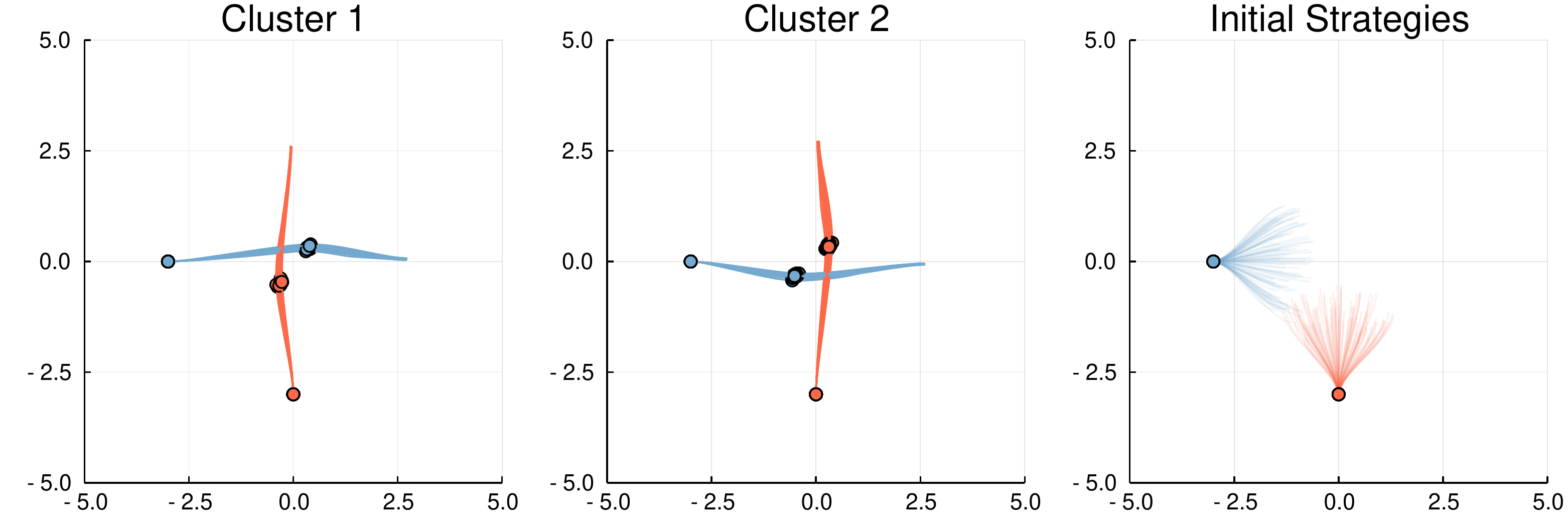}
  \caption{Clustered local equilibria for a 2-player navigation problem. Left: blue goes first. Center: red goes first. Right: initial strategies.}
  \label{fig:2-player-clustered}
\end{figure}

\Cref{fig:2-player-clustered} shows the two clusters of solutions that can be found in this setting.
Here, Player-1's trajectory is shown in blue, while Player-2's trajectory is shown in red.
The saturation of the respective colors indicates the average cost incurred by the player for the particular equilibrium, where higher saturation corresponds to lower cost.
Additionally, the clusters are sorted from left to right in ascending order of cost incurred by Player-1.
To give an intuition for time, for each trajectory we highlight each players initial position as well as their position when they are closest to the other player with circular marks.
This simple example shows that the game allows for two qualitatively different equilibria.
In the equilibrium corresponding to the first cluster Player-1 (blue) accelerates while Player-2 (red) slowly approaches the conflict area to let the other player pass first.
Additionally, both players slightly deviate from the straight path to their respective goal to make room for the other player as they share responsibility for collision avoidance.
In the equilibrium corresponding to the second cluster the players take opposite roles.
The examination of the cost for each cluster reveals that the player passing the conflict area first incurs a slightly lower cost than the player who is forced to decelerate and wait at the intersection.
Thus, it is clear that neither equilibrium dominates the other.
Instead, each cluster corresponds to different plausible mode of interaction.


\subsubsection{Running Example: 3-Player Scenario}

\begin{figure*}
  \centering
  \includegraphics[width=0.7\linewidth]{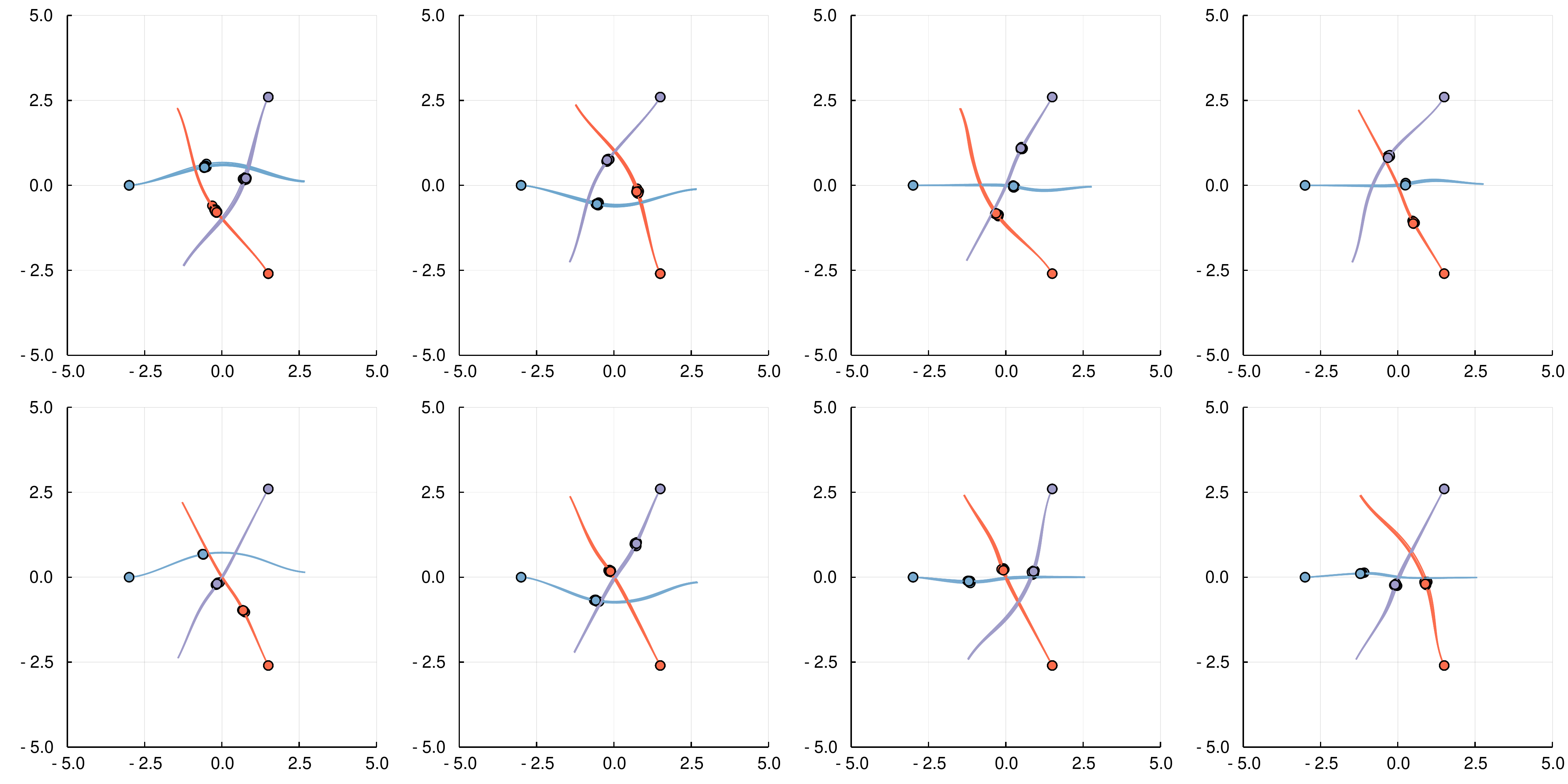}
  \caption{Clustered local equilibria for a 3-player navigation problem. Markers highlight positions at initial time and half the simulation horizon.}%
  \label{fig:3-player-unicycle-clustered}
\end{figure*}

We now add a third player to the navigation problem to recover our original running example as introduced in \cref{sec:problem}.
The clustered local equilibria for this scenario are depicted in \cref{fig:3-player-unicycle-clustered}.
Here, the third player's trajectories are shown in purple.
The clustering reveals a total number of eight qualitatively different local equilibria.
The two solutions with the lowest cost for all players ---~shown in the first two sub-figures of \cref{fig:3-player-unicycle-clustered}~--- correspond to a clockwise or counter-clockwise circular motion in which all players equally deviate from their straight path to the goal while maintaining almost constant speed.
The remaining solutions can be understood as all possible sequential orders in which players can pass the conflict zone.
That is, in each of these cases one player accelerates on the straight path to its goal to pass the conflict zone first.
Another player follows with moderate speed and slightly deviates from its straight path to the goal to avoid a collision.
The last player approaches the conflict zone at low speed and waits for the others to pass before continuing on a straight path to its goal.

The examination of qualitatively different approximate local Nash equilibria for the 3-player navigation scenario shows that the problem immediately becomes more complex, merely through the number of combinations in which the players may pass the conflict zone.
Therefore, it is clear that with increasing number of players hand-specifying suitable behaviors for all possible cases is undesirable and thus a more principled approach should be taken.

\subsection{Prediction}\label{sec:prediction-results}
In this section, we examine the performance of the proposed inference method independently from closed-loop interaction dynamics.
For this purpose, we again consider our running example of a 3-player navigation problem.
However, in this experiment the agent does not control any inputs to the system but rather observes the interaction of three players as they follow a set of strategies corresponding to a local equilibrium.
At every time step, the agent receives an observation of the physical state $x$.
With this information, it is the agent's task to accurately predict the trajectories of all players over a receding horizon of $T_p = \SI{10}{\second}$.
We compare our inference method proposed in \cref{sec:inference} to a baseline that does not actively reason about different solutions but instead uses a randomly sampled local equilibrium to predict the trajectories of all players.
Note that while the baseline does not actively reason about the equilibrium that the players operate at, it still benefits from state-feedback as it re-solves the game in a receding horizon fashion after every observation of the physical state.

\begin{figure}
  \centering
  \includegraphics[width=0.8\linewidth]{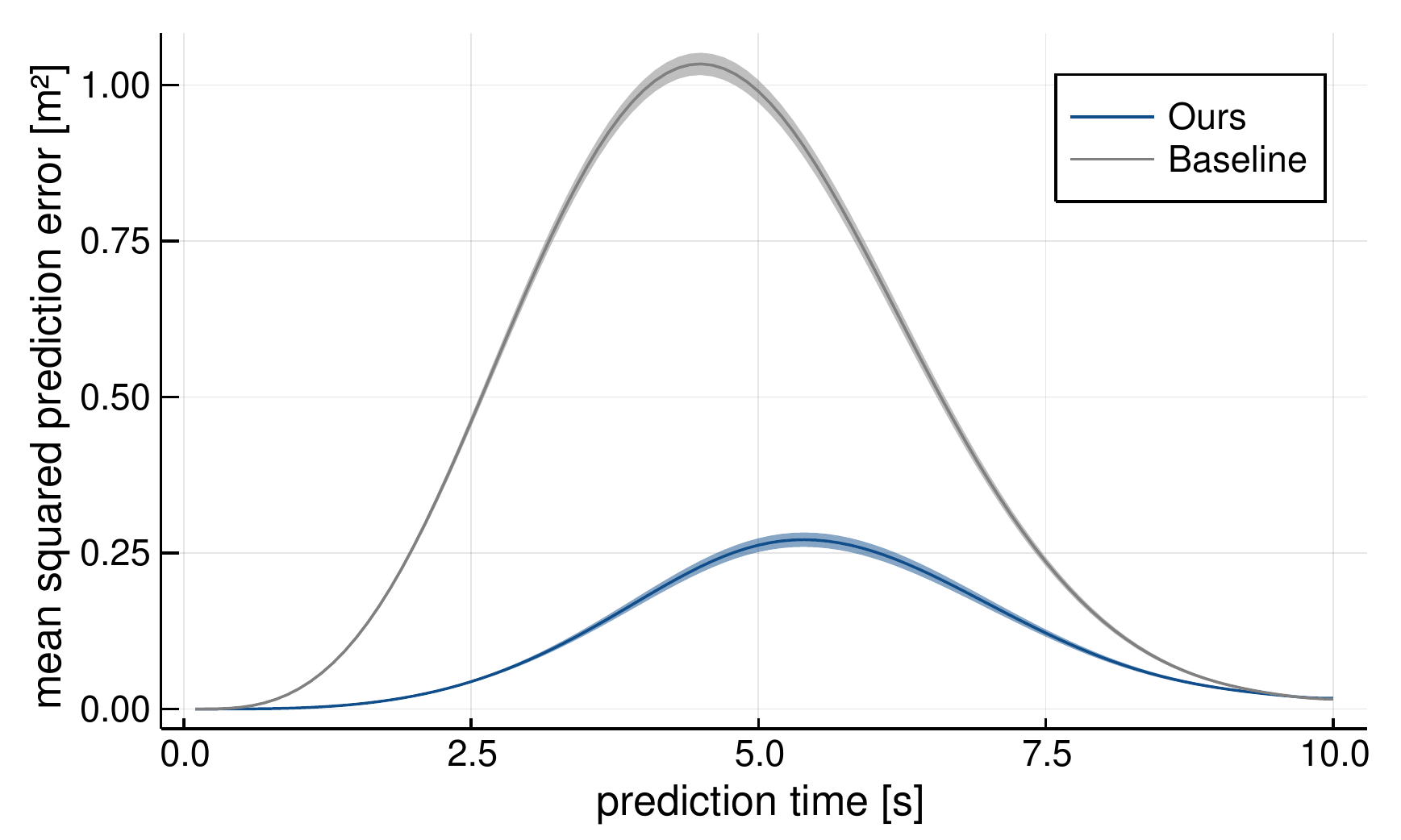}
  \caption{Mean squared prediction error over receding prediction horizon for the 3-player running example. Blue: Prediction with equilibrium inference (ours). Gray: Prediction with random local equilibrium (baseline).}%
  \label{fig:3-player-unicycle-prediction}
\end{figure}

\Cref{fig:3-player-unicycle-prediction} shows the mean squared prediction error $E\left[||p - \hat{p}||^2\right]$ of the position $p = [p_{x,1}; p_{y,1}; p_{x,2}; p_{y,2}; p_{x,3}; p_{y,3}]$ over the receding prediction horizon for our method and the baseline.
Each curve is generated from 100 simulations of the 3-player scenario.
Ribbons indicate the standard error of the mean.
As the agent receives an exact observation of the physical state at the current time and furthermore knows the goal locations $x_{g,i}$ of all players, both methods achieve a low prediction error at the beginning and the end of the prediction horizon.
However, in the intermediate range that is crucial to predict how conflicts are resolved our approach is able to significantly reduce the prediction error by inferring the equilibrium the game is played in.

\subsection{Planning With Strategy Alignment}\label{sec:planning-results}

Finally, we test the performance of the full inference and planning framework in a closed-loop interaction scenario.
In this experiment, the behavior for all human players is generated by solving the game for a randomly selected equilibrium.
All players generate their respective strategies over a receding horizon, therefore replanning after every time step to close the loop.
We compare our maximum a-posteriori aligned planner to a baseline that solves for a fixed equilibrium which we randomly sample at the beginning of each simulation run.

\subsubsection{Running Example: 3-Player Scenario}

\begin{figure}
  \centering
  \includegraphics[width=0.8\linewidth]{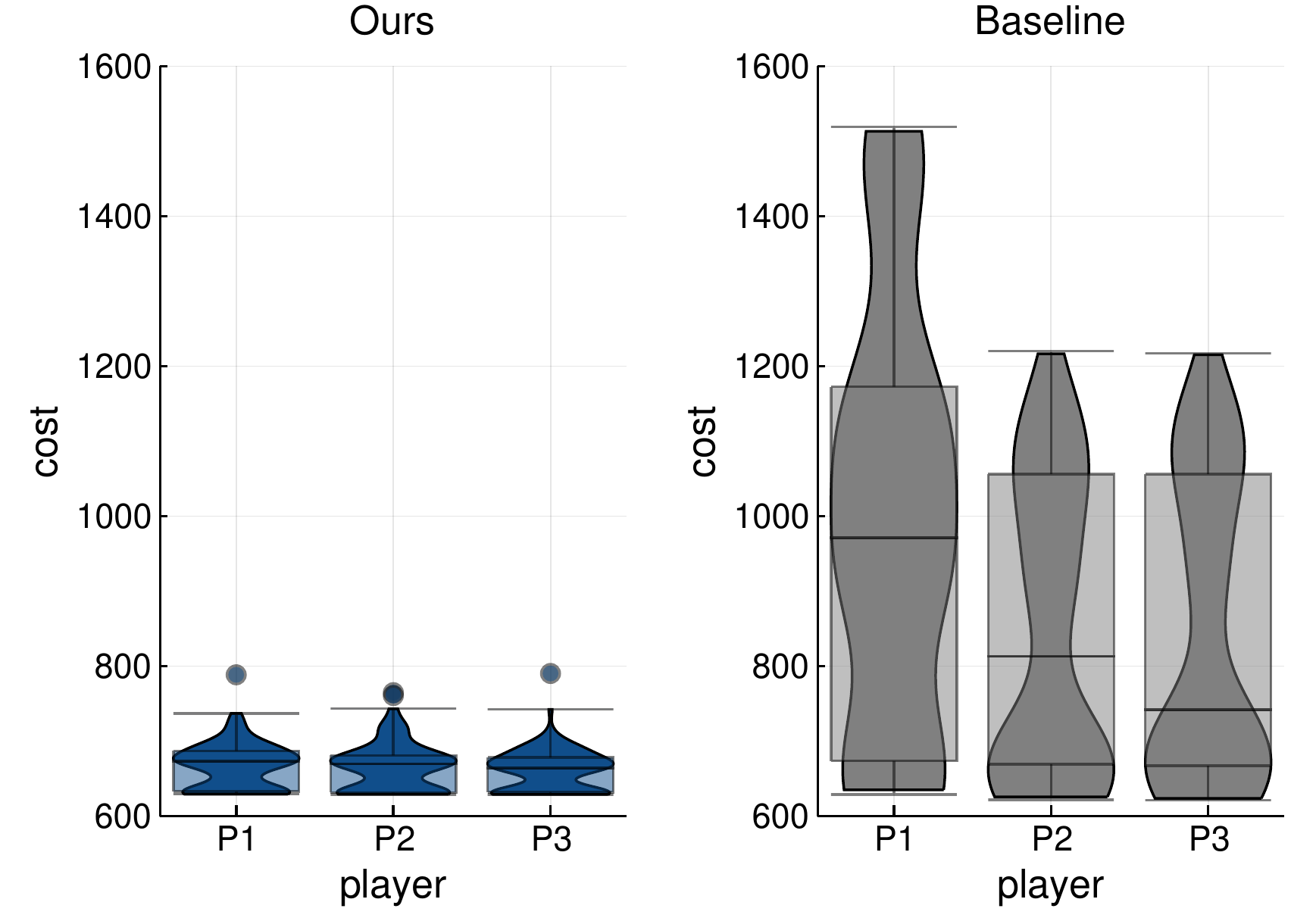}
  \caption{Distribution of costs incurred by each player. Left:  Player-1 uses strategy alignment (ours). Right: Player-1 operates at a random local equilibrium (baseline).}%
  \label{fig:3-player-unicycle-planning}
\end{figure}

\Cref{fig:3-player-unicycle-planning} shows the distribution of costs incurred by each player over 100 simulations of the 3-player scenario for both planning approaches.
It is clear that the maximum a-posteriori aligned planner performs significantly better than the non-adaptive baseline.
By actively aligning to the equilibrium chosen by the human players, the robot not only reduces its own cost but also the cost the remaining players.
In other words, inferring the intentions of the humans allow the robot to act more predictably and therefore permits a more favourable outcome due to the general sum nature of the game.

Moreover, we are able to solve this 3-player version of the strategy alignment problem at near real-time planning rates.
While solving for a single approximate local Nash equilibrium from scratch takes on average \SI{50}{\milli\second} for this example,
warm-starting allows to update solutions with an average runtime of just under \SI{2}{\milli\second}.
Thus, after having initialized all particles at the first time-step our planner runs in real-time when used in a receding horizon fashion.

\subsubsection{Scalability: 5-Player Scenario}

\begin{figure}
  \twocolumn[{
  \centering
  \includegraphics[width=0.8\linewidth]{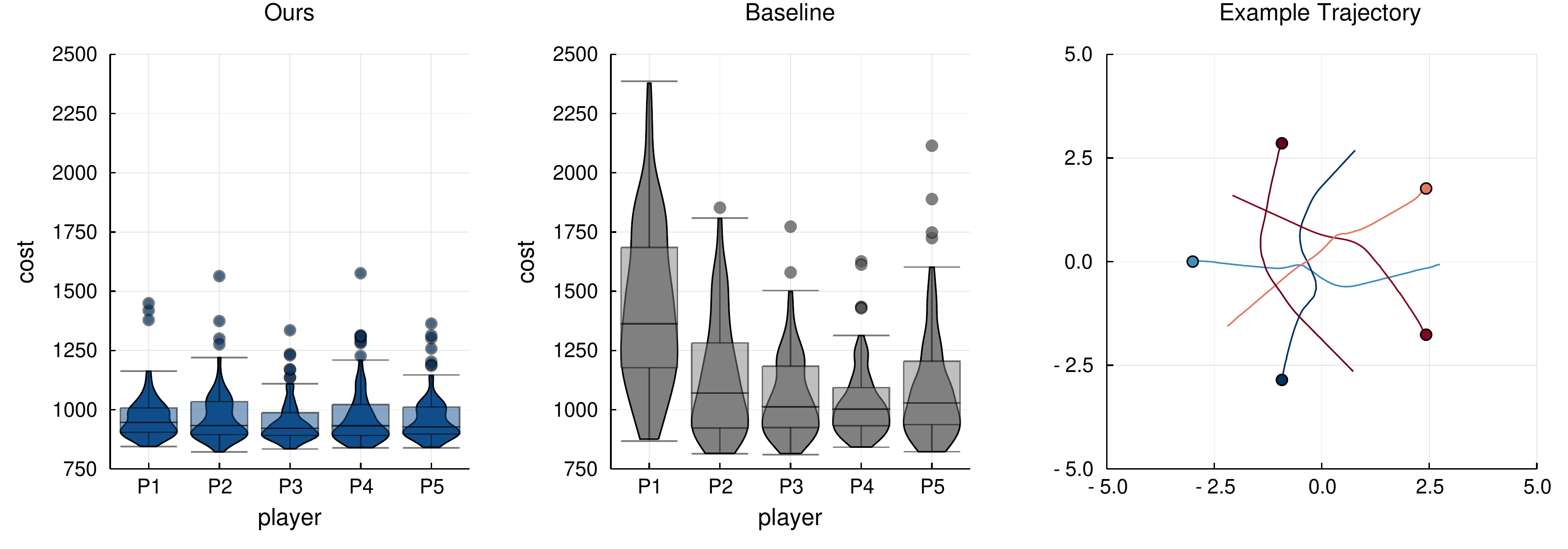}
  \caption{Distribution of costs incurred by each player in a 5-player navigation problem. Left: Player-1 uses strategy alignment (ours). Center: Player-1 operates at a random local equilibrium (baseline). Right: Example of a local equilibrium.\vspace{10pt}}%
  \label{fig:5-player-flat}
  }]
\end{figure}

Finally, we test scalability of our method by applying it to a 5-player version of the navigation problem.
In order to handle the increased complexity of the game, here we solve it by exploiting feedback linearization of the dynamics as proposed in \cite{fridovich2019flat}.
Note that for this purpose we reformulate the objectives directly in the linear coordinates.
Thus, the numerical value of the cost does not directly compare the costs received in the 3-player version of the game.

\Cref{fig:5-player-flat} shows the distribution of costs incurred by each player over 100 simulations of the 5-player scenario for our approach as well as the baseline.
Again, we observe a significant improvement by inferring the local equilibrium and aligning the robot's strategy accordingly.
In particular for the ego-agent this improvement is even more significant than for the 3-player running example as the increased complexity of the problem reduces the chance of choosing a compatible strategy without actively reasoning about the equilibrium targeted by other players.

It must be noted that the increased number of particles and the higher computational complexity of finding equilibria significantly increase the run-time for this problem.
In our current implementation of the 5-player navigation problem, solving for a single local equilibrium from scratch takes on average \SI{285}{\milli\second}, and \SI{131}{\milli\second} when utilizing warm-starting.
Accordingly, a planning step takes \SI{19.7}{\second} when updating all 150 particles using a single thread.
However, as many particles converge to the same equilibrium and particles are eliminated as the estimator becomes more confident the average runtime is only \SI{3.2}{\second} for our example.
Further improvements can be made by utilizing multiple threads for the belief update.

\section{Discussion \& Conclusion}

We have presented a general framework for game theoretic planning for problems in which the multiplicity of equilibria causes uncertainty about how agents are likely to behave.
We formulate the equilibrium that other players aim for as a hidden state that the agent can infer by contrasting their decisions to sampled approximate local Nash strategies in a technique similar to particle filtering.
By extracting the most likely hidden equilibrium we arrive at a planning scheme that allows the robot to align its predictions and its strategy with observed behavior.

We demonstrate our framework in simulations of a multi-player human-robot navigation problem which we formulate as a general-sum differential game.
Using this example, we show that by inferring which equilibrium humans are operating at, the robot is able to predict trajectories of humans more accurately, and observe that by aligning itself to this equilibrium, the robot is able to reduce the cost for \emph{all} players.
Due to the sample based belief representation and hinging on recent progress towards efficiently solving general-sum differential games to local equilibria \cite{fridovich2019efficient, fridovich2019flat}, we are able to run our method in real time for a 3-player navigation problem and demonstrate that even for 5 players the runtime is moderate.

In future work, we would like to test our framework in a human user-study to understand if our model applies to real world scenarios.
Beyond that, currently we plan with only the most likely equilibrium from the maintained belief.
As our method computes a particle belief over equilibria and corresponding feedback strategies for all players, a wide range of approximate planning techniques for partially observed decision making problems may be used to make better decisions in the face of high uncertainty.
The first of these techniques that we plan to explore is generalized QMDP.
Furthermore, our framework can be adapted to incorporate knowledge about how other players may change their choice of the desired equilibrium by modelling this decision as a probabilistic transition of the latent state.
In particular, this approach can be used to account for the fact that humans may be likely to change their strategy if otherwise they incur a high cost.
Finally, we would like to employ the prosed inference method to infer other latent variables such as parameters of the objectives of other players.


\bibliographystyle{ACM-Reference-Format}  
\bibliography{main.bib}  

\end{document}